\documentclass{amia}
\usepackage{lipsum} %Remove if not needed

\begin{document}

\title{Disparate Model Performance and Stability in Machine Learning Clinical Support for Diabetes and Heart Diseases}

\author{Ioannis Bilionis, MSc$^{1,2}$, Ricardo C. Berrios$^1$, Luis Fernandez-Luque, PhD$^1$, \\
Carlos Castillo, PhD$^{2,3}$}

\institutes{
    $^1$Adhera Health, Santa Cruz, USA; $^2$Universitat Pompeu Fabra, Barcelona, Spain; \\ $^3$ICREA, Catalonia, Spain
}

\maketitle

\section*{Abstract}

\textit{Machine Learning (ML) algorithms are vital for supporting clinical decision-making in biomedical informatics.  
However, their predictive performance can vary across demographic groups, often due to the underrepresentation of historically marginalized populations in training datasets.
% Goal 
The investigation reveals widespread sex- and age-related inequities in chronic disease datasets and their derived ML models. 
% Methods 
Thus, a novel analytical framework is introduced, combining systematic arbitrariness with traditional metrics like accuracy and data complexity.
% Results 
The analysis of data from over 25,000 individuals with chronic diseases revealed mild sex-related disparities, favoring predictive accuracy for males, and significant age-related differences, with better accuracy for younger patients. 
Notably, older patients showed inconsistent predictive accuracy across seven datasets, linked to higher data complexity and lower model performance.
% Conclusions 
This highlights that representativeness in training data alone does not guarantee equitable outcomes, and model arbitrariness must be addressed before deploying models in clinical settings.}

\section*{Introduction}
\label{sec:introduction}
 
% *** Motivation *** % To be given a more medical-based motivational aspect
%
In the realm of biomedicine, Artificial Intelligence (AI) methodologies, particularly Machine Learning (ML) models, are used as clinical support tools to systematically discern patterns and interdependencies among factors and outcomes within large datasets. 
ML has the potential to enhance healthcare provision by complementing, rather than supplanting, clinical judgment. It has demonstrated efficacy in the detection of skin cancer and diabetic retinopathy, among many other medical conditions~\cite{fogel2018artificial,jusman2021performance}.
A paramount objective when deploying ML models is the assurance of health equity~\cite{lo2023fundamentals,hamam2024guest}; thus, researchers and practitioners typically aim at attaining uniform model efficacy across diverse patient demographics~\cite{liu2023translational}.
The academic literature recommends an array of analytical tools for detecting biases, e.g., determining statistical dependencies between model outcomes, model errors, and specific subgroups~\cite{fletcher2021addressing}, particularly those experiencing both historical and ongoing discrimination.
Disparities detected in model performance are frequently attributed to deficiencies within the training datasets, typically lack of sufficient samples from those groups~\cite{buolamwini2018gender}.

% *** Related Work (General) *** %
%
Algorithmic fairness, as a research field, studies how and to which extent algorithmic decision support systems can be free from discriminatory biases~\cite{hajian2016algorithmic,raza2023connecting}.
Discrimination, in this context, means systematic disadvantages affecting socially salient groups~\cite{lippert2013born}. 
These disadvantages arise from a complex combination of design choices made at different points in the construction of an ML processing pipeline. %, from data collection and pre-processing, to model training and evaluation.
Discriminatory biases have been %extensively
documented in basically all applications of ML and AI~\cite{feuerriegel2020fair}, including recruitment~\cite{dastin2022amazon}, machine translation~\cite{caliskan2017semantics} and face recognition~\cite{hardesty2018study}, just to name a few.

% *** Related Work (Healthcare) *** %
%
In healthcare applications, prior research has identified algorithmic bias as a factor contributing to health disparities, highlighting the need for including Social Determinants of Health (SDoH) in ML to achieve health equity~\cite{tsai2022algorithmic, halamka2022addressing}.
For instance, in computer vision applications for medical imaging, biased data has been found to be a source of disparities in algorithmic outcomes~\cite{drukker2023toward,larrazabal2020gender}.
Differences in mortality prediction and X-ray diagnosis have been identified across racial/ethnic groups \cite{seyyed2020chexclusion, chen2019can},
including discrepancies in burn identification and diabetic retinopathy identification in dark-skinned versus lighter-skinned patients \cite{burlina2021addressing,abubakar2020assessment}, and in an opioid misuse classifier, with more errors (false negatives) for dark-skinned patients \cite{thompson2021bias}.
In other cases, ML algorithms have predicted similar risk scores in both light- and dark-skinned patients, even though the dark-skinned patients had higher risk~\cite{obermeyer2019dissecting, park2021comparison}.
There are many other examples, as this is an active research topic that to some extent is in its early stages~\cite{jobin2019global, marabelli2021lifecycle, leavy2018gender,anderson2024measuring}.
% \clearpage % TEMPORARY

% *** Methodology (traditional) *** %
An in-depth knowledge of an ML application and of its context should inform this analysis~\cite{suresh2021framework}. 
In healthcare, the generalizability of AI algorithms across subgroups is critically dependent on training datasets, including factors such as representativeness, missing data, and outliers~\cite{ahmad2020fairness}.
This suggests that some biases can be traced to datasets that underrepresent certain populations; using these unbalanced datasets as training data yields algorithmic models that exhibit systematically unbalanced errors~\cite{corbett2017algorithmic}.
In this context, the augmentation of the dataset with additional samples from the underrepresented group, which frequently corresponds to groups that are socioeconomically disadvantaged or medically underserved, has been empirically demonstrated to mitigate discrepancy in model accuracy. This is the case of the seminal ``Gender Shades'' study \cite{buolamwini2018gender, raji2019actionable}. Similar results have been observed in the training set of a popular face detection benchmark dataset \cite{yang2022enhancing}.

% *** Methodology (ours) *** %
Differences in algorithmic performance are not always due to lack of representativeness.
Signs and symptoms of many conditions vary between different populations~\cite{boehme2014racial, cirillo2020sex,celeste2023ethnic,hou2024pferm}.
Crucially, the features included in a dataset may be more or less useful for predicting different outcomes (e.g., being clinically diagnosed with a condition or not). 
The analysis of a dataset under this perspective is known as \emph{data complexity} analysis, and it encompasses multiple aspects. A significant body of research has been dedicated to the formulation of various metrics that encapsulate the multifaceted aspects of dataset complexity~\cite{lorena2019complex}.
Beyond disparities in model accuracy and data complexity, recent work highlights the importance of variance in model predictions. This variance is related to the extent to which model predictions can ``flip'' under minor changes in the training data, and it becomes an aspect of algorithmic fairness when high-variance predictions are concentrated in a demographic subgroup. This is called \emph{systematic arbitrariness}~\cite{cooper2024arbitrariness}

% *** Contribution *** %
This paper describes a multifaceted analysis of training datasets pertinent to chronic diseases aimed at uncovering potential discrepancies that could lead to biases in the resulting ML models.
Our research substantiates the premise that demographic parity within datasets does not inherently ensure uniformity in algorithmic performance.
That is to say, even datasets that are ostensibly equitable in terms of demographic attributes may still yield models with performance discrepancies.
Initiating our analysis with a common ML performance metric, the Area Under the Receiving Operating Characteristic curve (AUROC or AUC), we measure the predictive efficacy of the models. 
Subsequently, our examination extends to more profound dataset attributes impacting model behavior, particularly data complexity and systematic arbitrariness. 
Our methodology provides a comprehensive and systematic framework for assessing training data from the perspective of algorithmic fairness, incorporating three key steps: (1) Model Performance Analysis, which evaluates disparities in predictive accuracy between demographic groups using robust cross-validation and bootstrapping techniques; (2) Data Complexity Metrics, leveraging established metrics to explore the intrinsic challenges of the datasets, such as feature informativeness and class imbalance; and (3) Systematic Arbitrariness Examination, quantifying inconsistencies in model predictions and their concentration across specific data subsets. To the best of our knowledge, this investigation is the first to test systematic model arbitrariness in the healthcare domain.

\section*{Methods}
\label{sec:methods}
% *** Dataset *** %
\subsection*{\textit{Datasets}}
% Dataset's description
We use a list of datasets identified and reported in a survey of publicly accessible datasets related to chronic diseases ~\cite{bilionis2023survey}.
Within this selection, two datasets pertain to  diabetes ($D_1,D_2$), while five are related to cardiac conditions ($D_3 \dots D_7$).
Dataset sizes vary widely (see Table~\ref{tab:dataset_descr}), and for the purpose of this study,  we segmented  two of the large datasets into smaller subsets ($D_{2a}, D_{2b}, D_{7a}, D_{7b}$) by randomly selecting two samples, each sized 100 times larger than the number of attributes.
For the purpose of analysis, sex and age variables are binarized. In the case of age, the individuals within the lowest two quintiles are categorized as ``young'', and those within the highest two quintiles are categorized as ``old'',  with the median quintile remaining unassigned.
Dataset $D_1$ does not include sex. Dataset $D_{7\{a,b\}}$ was made available in 2020, but the specific  year of data collection is not explicitly documented.

\begin{table}[h]
    \caption{Characteristics of the datasets used in this research.}
    \centering
    \small % Further reduce the font size
    % \setlength{\tabcolsep}{1pt} % Adjust column padding
    % \begin{tabular}{p{0.5cm}p{0.5cm}p{0.5cm}p{0.5cm}p{0.5cm}p{1.5cm}p{1.5cm}}
    \begin{tabular}{@{}ccccccc@{}}
        \toprule   
        \begin{tabular}[c]{@{}c@{}}\text{Dataset} \\ \text{ID}\end{tabular}  &
        \begin{tabular}[c]{@{}c@{}}\text{Therapeutic} \\ \text{Area}\end{tabular} &
        \multicolumn{1}{c}{\text{N}} &
        \multicolumn{1}{c}{\text{Year}} &
        \begin{tabular}[c]{@{}c@{}}\text{Sex Ratio} \\ \text{Female:Male}\end{tabular} &
        \begin{tabular}[c]{@{}c@{}}\text{Younger Group} \\ \text{Age Range}\end{tabular} &
        \begin{tabular}[c]{@{}c@{}}\text{Elder Group} \\ \text{Age Range}\end{tabular} \\
        \midrule
        \text{\(D_{1}\)}  & \text{Diabetes} & 768 & 1988 & - & [21, 23] & [33, 81] \\
        \text{\(D_{2a}\)}  & \text{Diabetes} & 4,400 & 2014 & 1.17 & [5, 65] & [75, 95] \\
        \text{\(D_{2b}\)} & \text{Diabetes} & 4,400 & 2014 & 1.09 & [5, 65] & [75, 95]  \\
        \text{\(D_{3}\)}  & \text{Heart Dis.} & 920 & 1989 & 0.29 & [28, 52] & [57, 77] \\
        \text{\(D_{4}\)} & \text{Heart Dis.} & 452 & 1997& 1.27 & [0, 43] & [51, 83] \\
        \text{\(D_{5}\)} & \text{Heart Dis.} & 4,240 & 2010 & 1.33 & [32, 46] & [52, 70] \\
        \text{\(D_{6}\)}  & \text{Heart Dis.} & 10,000 & 2020 & 0.79 & [4, 57] & [66, 98] \\
        \text{\(D_{7a}\)}  & \text{Heart Dis.} & 1,300 & \textit{ca.}2020 & 1.86 & [30, 52] & [56, 65]  \\
        \text{\(D_{7b}\)}  & \text{Heart Dis.} & 1,300 & \textit{ca.}2020 & 1.93 & [30, 52]  & [56, 65] \\\bottomrule
    \end{tabular}
    \label{tab:dataset_descr}
\end{table}

% *** Model Performance Disparities *** %
\subsection*{\textit{Model Performance}}

For the evaluation of model performance, we used three gradient boosting algorithms (XGBoost~\cite{chen2016xgboost}, LGBoost~\cite{ke2017lightgbm}, HGBoost~\cite{guryanov2019histogram}) that support missing values.
We considered two sets of attributes: including the protected attributes (``aware model''), and excluding them (``unaware model''). The performance metrics were  similar across both models, which means that the datasets contain proxy variables for the protected attributes.
Model training was done using a 3-fold cross validation schema, which involves partitioning the dataset into three subsets and cyclically using two-thirds for training and one-third for testing.
This evaluation was further complemented by repeated bootstrapping, wherein each iteration involved a novel partitioning of the dataset. 
Hence, each reported Area Under ROC Curve value (AUROC, or simply ROC) is the average of 66 models: 3 algorithms times 22 runs (19 random runs plus 3 cross-validation runs).

\begin{figure}[ht]
    \centering
    \subfloat[Learning curves]{
        \includegraphics[width=0.60\textwidth]{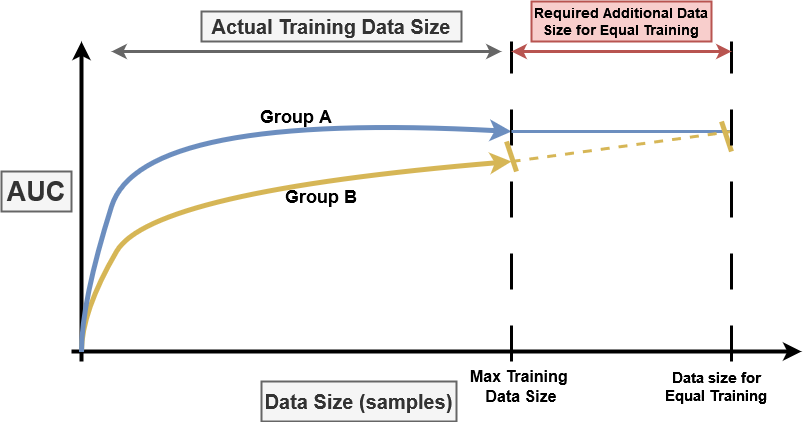}
        \label{subfig:curve_disparities}
    }\hfill\subfloat[Self-Consistency]{
        \includegraphics[width=0.36\textwidth]{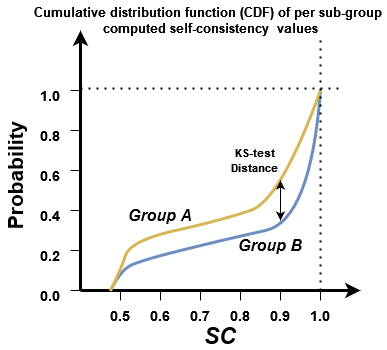}
        \label{subfig:consistency_disparities}
    }
    \caption{Depiction of our methods regarding learning curves and self-consistency.}
    \label{fig:auc_sc_disparities}
\end{figure}

Learning curves were obtained through an analogous process.
We extrapolated learning curves to deduce an estimate of the number of additional data points that would enable the group with lower performance to attain the benchmark set by the group with higher performance. 
Figure~\ref{subfig:curve_disparities} illustrates our method. Let $f(n)$ be the superior learning curve, and $g(n)$ be the inferior learning curve, with $h(n)$ being an extrapolation of the learning curve.
Conservatively, if we assume the upper curve reaches a saturation point $f(N_p) = AUC_p$ (which is not always the case, hence the conservative estimate), we attempt the following minimization:
\begin{align*} \label{eq:min_additional_points}
\min \quad & N_{add} \\
\textrm{s.t.} \quad & h(N_p + N_{add}) = f(N_p) 
\end{align*}

\noindent i.e., we calculate the minimal number of additional data points that would be required from the group with the lower AUC to match the AUC of the group with higher performance. 
We consider three different functions $h_1(\cdot), h_2(\cdot), h_3(\cdot)$, each linearly constructed based on different segments of the concluding portion of the learning curve, choosing the one that necessitates the smallest increase in data points (i.e., the most conservative scenario, to avoid exaggerating the discrepancy). 
The greater the value of  $N_{add}$, the larger the performance disparity.

% *** Data Complexity *** %
\subsection*{\textit{Data Complexity Metrics}}
\label{subsec:methods-complexity}

Data complexity analysis is a systematic effort to understand discrepancies in classification accuracy by relating them to intrinsic characteristics of a dataset.
This is a large research topic, and the interested reader can consult any of various surveys about it~\cite{santos2023unifying,rivolli2022meta,gupta2021data}.
We used a fairly standard categorization of data complexity metrics~\cite{lorena2019complex}, and picked one popular complexity metric within each category, as shown in~Table~\ref{tab:data-complexity}.

\begin{table}[ht]
    \caption{Data complexity: metrics categories and representative used; see \cite{lorena2019complex} for details.}
    \centering
    \small
    % \setlength{\tabcolsep}{3pt} % Adjust column separation
    % \begin{tabular}{p{0.25\columnwidth}p{0.35\columnwidth}p{0.3\columnwidth}}
    \begin{tabular}{lll}
    \toprule
    Family name & Object of analysis & Metric used \\
    \midrule
    Feature-based & Feature informativeness &\begin{tabular}[l]{@{}l@{}}\text{ Max. Fisher’s} \\ \text{discriminant ratio}\end{tabular}   \\
    \addlinespace
    Linearity & Linear separability & \begin{tabular}[l]{@{}l@{}}\text{Sum of the error distance} \\ \text{by linear programming}\end{tabular} \\ 
    \addlinespace
    Neighborhood & Local class distribution & \begin{tabular}[l]{@{}l@{}}\text{Error rate of } \\ \text{nearest-neighbors classifier}\end{tabular}\\
    \addlinespace
%    Network & Graph structure details & \inote{???} \\
    Dimensionality & Data dimensionality/sparsity & \begin{tabular}[l]{@{}l@{}}\text{Average number of} \\ \text{features per point}\end{tabular} \\
    \addlinespace
    Class imbalance & Ratio between class examples & Imbalance ratio\\
    \bottomrule
    \end{tabular}
    \label{tab:data-complexity}
\end{table}

% *** Systematic Arbitrariness *** %
\subsection*{\textit{Systematic Arbitrariness}}

A family of models (e.g., various models built using the same learning scheme but different portions of the training data) may exhibit arbitrariness.
Model arbitrariness corresponds to discrepancies in the predicted label for some elements across models of the same family, and it tends to be systematic, i.e., concentrated on specific items.

A recent study introduces a metric of Self-Consistency (SC)~\cite{cooper2024arbitrariness}, which is computed at the level of an item as the probability that two models of the same family agree on the label for an item.
For instance, an item with self-consistency of 1.0 is an item for which any model of a family yields the same predicted label. In binary classification, the minimum self-consistency is 0.5, indicating that half of the models yield one predicted label, and half of the models yield the opposite label.
Note that self-consistency is independent of the ``true'' label of an element.

To compare self-consistency scores between groups, as recommended in \cite{cooper2024arbitrariness} we use the Cumulative Distribution Function (CDF) of self-consistency. Often, one curve is above another, similarly to what we see in Figure~\ref{subfig:consistency_disparities}. 
We measure the disparity by performing a statistical test of the difference between the two curves.

\section*{Results}
\label{sec:results}

% *** Model Performance Disparities *** %
\subsection*{\textit{Variations in model performance}}
Using datasets collected in prior research, which include patient demographic details such as sex and age~\cite{bilionis2023survey}, our approach leverages three distinct gradient boosting algorithms to infer ML models from the training data. 
The validation methodology used herein incorporates cross- validation complemented by iterative bootstrapping, thereby generating a multitude of models each informed by different subsets of the training data. 
By examining the Area Under the ROC Curve (AUC), we determine the model’s proficiency in distinguishing between patient cohorts with different clinical outcomes.
Table~\ref{tab:auc-differences} presents a disaggregated view of AUC discrepancies across sex and age demographics.
These results account for models that incorporate age and sex as predictive attributes (“aware modeling”). Similar results are observed when these variables are omitted (“unaware modeling”). 

\begin{table*}[ht]
    \caption{Model performance (AUC), with average and variance computed over 66 models for each cell in the table. Differences along sex/age are expressed using p-values: $<0.01$ (*), $<0.001$ (**), $<0.0001$ (***). The highest AUC is in boldface when the difference is significant at $p<0.01$.}
    \centering
    \small
    \begin{tabular}{cccc|ccc}
    \toprule
    \multirow{3}{*}{Dataset} & \multicolumn{6}{c}{AUC} \\\cline{2-7}
     & Female & Male & p-value 
     & Old & Young & p-value \\\midrule
     $D_1$ & -- & -- & -- & 0.65 $\pm$ 0.04 & \textbf{0.69} $\pm$ 0.05 & $<0.0001$ *** \\
     $D_{2a}$ & 0.59 $\pm$ 0.02 & \textbf{0.61} $\pm$ 0.02 & $<0.0001$ *** 
       & 0.58 $\pm$ 0.02 & \textbf{0.62} $\pm$ 0.02 &$<0.0001$ *** \\
     $D_{2b}$ & 0.59 $\pm$ 0.02 & 0.59 $\pm$ 0.02 & 0.67 
       & 0.58 $\pm$ 0.02 & \textbf{0.60} $\pm$ 0.02 & $<0.0001$ *** \\
     $D_3$ & 0.66 $\pm$ 0.07 & \textbf{0.71} $\pm$ 0.06 & $<0.0001$ *** 
       & 0.63 $\pm$ 0.06 & \textbf{0.73} $\pm$ 0.06 &$<0.0001$ *** \\
     $D_4$ & 0.77 $\pm$ 0.04 & 0.78 $\pm$ 0.06 & 0.80 
       & 0.79 $\pm$ 0.05 & 0.79 $\pm$ 0.06 & 0.95 \\
     $D_5$ & 0.53 $\pm$ 0.02 & \textbf{0.56} $\pm$ 0.02 & $<0.0001$ *** 
       & \textbf{0.54} $\pm$ 0.02 & 0.50 $\pm$ 0.01 & $<0.0001$ *** \\
     $D_6$ & \textbf{0.92} $\pm$ 0.01 & 0.87 $\pm$ 0.02 & $<0.0001$ *** 
       & 0.87 $\pm$ 0.02 & \textbf{0.91} $\pm$ 0.01 & $<0.0001$ *** \\
     $D_{7a}$ & \textbf{0.68} $\pm$ 0.03 & 0.67 $\pm$ 0.03 & $<0.01$ * 
       & 0.62 $\pm$ 0.03 & \textbf{0.69} $\pm$ 0.03 & $<0.0001$ *** \\
     $D_{7b}$ & \textbf{0.71} $\pm$ 0.02 & 0.67 $\pm$ 0.03 & $<0.0001$ *** 
       & 0.66 $\pm$ 0.02 & \textbf{0.70} $\pm$ 0.03 & $<0.0001$ *** \\
       \bottomrule
    \end{tabular}
    \label{tab:auc-differences}
\end{table*}

In our analysis, we observe disparities between sexes within several models, and across age groups in all but one model.
Regarding sex-based disparities, approximately 10\%  of validation results reveal a higher AUC for males compared to females, wheras a mere 1\% of results show higher female AUC relative to male AUC.
Regarding age-related variances, these disparities exceed the sex-related ones with 32\% of validation results demonstrating that the AUC for younger patients exceeds that of older patients, and conversely, in 5\% of the cases, the AUC is greater for older patients compared to the younger patients.

% *** Model Training Disparities *** %
\subsection*{\textit{Learning curves and the expected impact of additional data}}
Learning curves are a standard tool for monitoring changes in model performance with the incremental addition of training data points. 
These curves graphically depict a performance metric, such as AUC, against the volume of  training data utilized to build the model. 
Through this visual representation, one can appreciate trends like the plateauing of performance gains and extrapolate the requisite quantity of additional data points necessary to attain a predefined AUC level in scenarios where the learning curve does not plateau. 

Consistently aligning with our previous results regarding the AUC obtained from the comprehensive training datasets (excluding the fraction set aside for testing during cross-validation), it is often observed that the learning curve for one demographic group is above the learning curve for the other. 
This phenomenon suggests that even with balanced training sets, the resultant AUC may favor one group over another.
Such a trend provides indirect evidence of differences in the predictability of outcomes between two groups when identical training data volumes and features are employed. 

We have estimated the number of additional data points from the group exhibiting lower performance that would be required to match the AUC of the group with superior performance, based on the extrapolation of the learning curves to the point of anticipated AUC parity, given the current trajectory. 
Table~\ref{tab:learning-curves-summary} summarizes our results.
The symbol Infinity ($\infty$) means that the learning curve for the group for which the model has lower performance appears to plateau at an AUC threshold, signifying no further enhancement with additional data points.

Our analysis reveals that there are often imbalances that, to be corrected, would require a substantive amount of additional training instances. %; in certain scenarios, this could mean augmenting the existing dataset by up to 100\%.
In half of the datasets, equating the AUC for sex would require the addition of 2\% to 57\% additional data for females, while for the remainder, an increment of 3\% and 48\% would be required for males. 
Regarding age, achieving parity would involve acquiring 2\% and 46\% more data for the younger patients in two datasets, and a substantial 5\% and 192\% increase for the older patients in the other datasets.

\begin{table}[ht]
    \caption{Estimation, obtained by extrapolating learning curves, of the additional data points ($N_{add}$) needed to achieve AUC parity.}
    \centering
    \small
    \begin{tabular}{ccccc}
    \toprule
    Dataset & Group & $N_{add}/N$ & Group & $N_{add}/N$ \\\midrule
    $D_1$ & -- & -- & Old & 192\% \\
    $D_{2a}$ & Female & 13\% & Old & 112\% \\
    $D_{2b}$ & Female & 2\% & Old & 129\% \\
    $D_3$ & Female & 66\% & Old &  $\infty$ \\
    $D_4$ & Male & 3\% & Young & 2\% \\
    $D_5$ & Female & 57\% & Young & 46\% \\
    $D_6$ & Male & 48\% & Old & 8\% \\
    $D_{7a}$ & Male & 6\% & Old & 5\% \\
    $D_{7b}$ & Male & 4\% & Old & 33\% \\
    \bottomrule
    \end{tabular}
    \label{tab:learning-curves-summary}
\end{table}

Focusing on datasets that would benefit from at least a 10\% increase in data, in 3 out of the 8 datasets where sex data is present, additional training data are required for females. In a similar vein, for age-related imbalances, 4 out of the 9 datasets would need additional data for the older patients group to achieve AUC parity.

% \clearpage % TEMPORARY

% *** Data Complexity *** %
\subsection*{\textit{Alignment of data complexity with some disparities}}

We considered sixteen complexity metrics grouped into five categories, each corresponding to a unique conceptual framework, and computed the disparity of each metric between protected subgroups regarding sex and age. 
For each data set, AUC disparity divided by complexity metric disparity creates a ratio reflecting the consistency between model performance and data complexity as follows (where CM: Complexity Metric and A,B: Sub-groups A and B, i.e. Female-Male and Old-Young):

\[
\frac{{\overline{\text{AUC}_A} - \overline{\text{AUC}_B}}}{{\text{CM}_B - \text{CM}_A}} = \begin{cases} 
1   \text{ (Consistency)} & \text{if } x > 0  \\ 
-1  \text{ (Inconsistency)} & \text{if } x \leq 0  
\end{cases}
\]

\begin{figure}[ht]
    \centering
    \subfloat[AUC-Complexity Consistencies by Sex]{
        \includegraphics[width=0.45\textwidth]{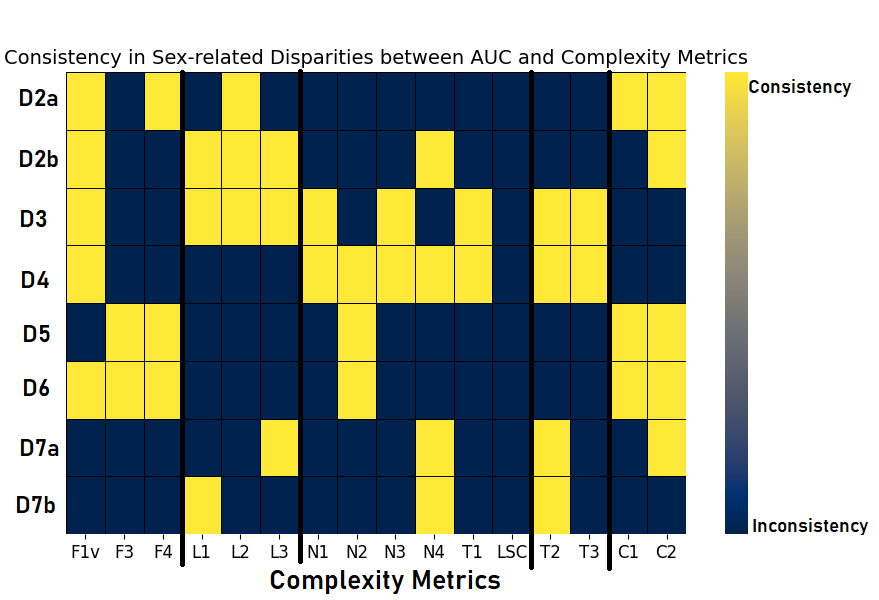}
        % \caption{a}
        \label{subfig:data_complex_sex}
    }
    \hfill
    \subfloat[AUC-Complexity Consistencies by Age]{
        \includegraphics[width=0.45\textwidth]{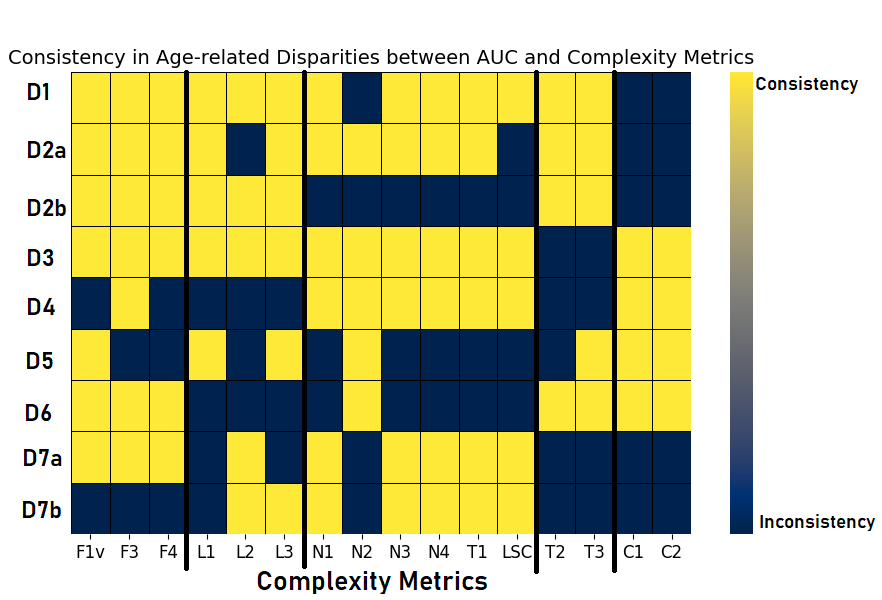}
        % \caption{b}
        \label{subfig:data_complex_age}
    }
    \caption{Consistency of sex- and age-related disparities between AUC and complexity metrics. Light colors indicate the cases in which auc and data complexity disparities are consistent, while dark colors inconsistent.}
    \label{fig:auc_complex_heatmap}
\end{figure}

Figure~\ref{fig:auc_complex_heatmap} presents the results in a heatmap visualization highlighting with light color the cases in which higher complexity and lower AUC values are observed for a specific sub-group in comparison with the other, while dark color indicates inconsistent AUC and complexity patterns.

No obvious patterns can be observed across data in the results. Indeed, there are some situations of complementarity in which complexity metrics that are well aligned with AUC in some datasets are not aligned with AUC for another dataset and vice versa.
Nevertheless, more complex data could potentially be linked to lower model performance, as homogeneous behavior is observed for some categories of metrics (especially in feature, dimensionality and class imbalance) within datasets regarding age.
In addition, several sets of databases (e.g. those related to diabetes $D_1,D_{2a},D_{2b}$) show consistent disparities between performance and i) feature-based and ii) dimensionality complexity metrics. 
However, these experiments suggest that complexity metrics cannot be relied upon as a predictor of AUC disparities in specific clinical conditions.
% \clearpage % TEMPORARY

% *** Systematic Arbitrariness *** %
\subsection*{\textit{Systematic arbitrariness and model stability}}

In analyzing a family of models, each trained on distinct yet equivalently-sized partitions of the training data, we define an individual’s self-consistency as the probability that two models within this family will yield the same label~\cite{cooper2024arbitrariness}. 
 In our case, the minimum self-consistency is attained by subjects for which half of the models predict that they will be diagnosed with a condition, while the other half predict that they will not.
Evidently, this is a situation we would like to avoid as much as possible.
Hence, all other things equal, a model with higher self-consistency for most items is preferable.

Systematic arbitrariness is observed when items with low self-consistency are concentrated within a particular group, and can be measured by comparing CDFs. 
To quantify disparities, we use the Kolmogorov-Smirnov (KS) statistical test (Figure~\ref{subfig:consistency_disparities}), with the results shown in Table~\ref{tab:self-consistency-results}.

\begin{table*}[ht]
    \caption{Sub-group arbitrariness: area under the CDF of self-consistency results for each sub-group and distance between them measured by Kolmogorov-Smirnov (KS) statistical test. The results of the significance test indicate p-value$<$0.01 (*), $<$0.001 (**), and when significant the larger arbitrariness appears in boldface.}
    \centering
    \small 
    \begin{tabular}{cc|cccc|cccc}
        \toprule
        &  & \multicolumn{4}{c|}{\textbf{Sex}} & \multicolumn{4}{c}{\textbf{Age}} \\
        Dataset & \textit{Overall} & \textit{Female} & \textit{Male} & \textit{KS-test} &
        & \textit{Elder} & \textit{Younger} & \textit{KS-test} & \\
        \midrule
        $D_{1}$&  0.243 & - & - & -             & & \textbf{0.247} & 0.172 & 0.205 & * \\
        $D_{2a}$& 0.323 & 0.308 & 0.312 & 0.019 & & 0.295 & 0.294 & 0.045  \\
        $D_{2b}$& 0.332 & 0.313 & 0.319 & 0.020 & & 0.302 & 0.299 & 0.018  \\
        $D_{3}$&  0.258 & 0.223 & 0.238 & 0.071 & & 0.222 & 0.210  & 0.043 \\
        $D_{4}$&  0.313 & 0.233 & 0.247 & 0.037 & & 0.279 & 0.263  & 0.057  \\
        $D_{5}$&  0.119 & 0.105 & 0.119 & 0.060 & & \textbf{0.169} & 0.069  & 0.299 & ** \\
        $D_{6}$&  0.084 & 0.083 & 0.078 & 0.031 & & 0.078 & 0.085 & 0.024 \\
        $D_{7a}$& 0.266 & 0.244 & 0.233 & 0.053 & & \textbf{0.263} & 0.233 & 0.082 & *  \\
        $D_{7b}$& 0.269 & 0.247 & 0.250 & 0.049 & & \textbf{0.262}  & 0.228 & 0.088 & * \\
        \bottomrule
    \end{tabular}
    \label{tab:self-consistency-results}
\end{table*}

Results show that ML predictions have no significant self-consistency disparities between male and female subjects. However, older individuals exhibit significantly more arbitrary predictions in 4 out of 9 datasets.
These results are for a model including protected characteristics (``unaware model''). Results for the aware model are similar, although with lower self-consistency values in general and smaller differences by sex and age.

% \clearpage % TEMPORARY

%%%%%%%%%%%%%%%%%%%%%%%%%%%%%%%%%%%%%%%%%%%%%%%%%%%%%%%%%%%%%%%%
\section*{Discussion}
\label{sec:discussion}

% *** Summarized Results *** %
Our findings uncover sex- and age-related disparities in model performance as evidenced by the AUC of the models. It is pertinent to recall that the representation of each group within the datasets is equal.
This observation underscores that mere demographic parity in training datasets does not mean model equity. 
The analysis of learning curves provides insights into the potential benefits of data augmentation. 
Sex-related disparities are observed to occasionally favor males over females, with a marginal predominance for male patients as indicated by both AUC and the requisite additional data to attain performance parity. 
Regarding age differences, the findings are more pronounced, with models generally predicting better for younger patients across most datasets (higher AUC), and requiring a large volume of additional training data to potentially achieve performance parity.

Furthermore, upon examining disparities in data complexity and systematic arbitrariness, we observe that predictions for older patients tend to be less consistent than those for their younger counterparts in several datasets. 
These disparities, to some extent, correlate with the model performance (AUC) and data complexity findings, suggesting a linkage between increased data arbitrariness for older patients and heightened complexity, leading to lower model performance. 
These correlations suggest but do not determine model disparities, as there are exceptions within our observations, where greater arbitrariness is sometimes associated with comparable or superior AUC values. 
This highlights the necessity of a multifaceted metric consideration encompassing performance, complexity, and stability, rather than relying exclusively on performance metrics.

% *** Implications for practice *** %
Within the healthcare domain, the legal and ethical dimensions of decision-making are of paramount importance~\cite{peters2020responsible}. 
The findings of this study highlight some characteristics of model performance that are not typically reported, but that hold considerable potential to influence clinical practice. 
Specifically, systematic arbitrariness in model outputs could undermine clinician confidence in ML and diminish the acceptability of such models. 
In nearly half of the datasets we studied, older patients with chronic diseases face the risk of health inequities~\cite{lorenc2012types, klang2020sex} due to data that is suboptimal for modeling their health outcomes as compared to younger patients. 
%
% We propose datasets are tested for systematic arbitrariness before being used in clinical settings.
To address these challenges, datasets and models should undergo rigorous testing for systematic arbitrariness prior to deployment, incorporating self-consistency metrics to detect and mitigate prediction discrepancies across demographic groups. 
Model developers must prioritize detailed fairness audits, including systematic arbitrariness and demographic-specific performance metrics, as part of standard validation protocols. 
Healthcare practitioners are integral to the ethical implementation of ML models by insisting on transparency from developers, carefully evaluating outputs for potential biases, and complementing ML recommendations with clinical expertise.
They should actively promote inclusivity by pinpointing data deficiencies that disadvantage certain groups and collaborating with developers to mitigate these gaps, while encouraging continuous model evaluation to refine performance over time.

% *** Limitations *** %
Hospital data, such as the one used in this study, may be indicative solely of the population with healthcare system access, thus potentially engendering bias against certain subpopulations~\cite{chen2019deep,hyun2017gender,gauci2022biology,veinot2018good}.
Future efforts should aim to extend these analyses to include additional databases.
% *** Future Work *** %
To address situations where systematic arbitrariness is detected, we must consider both technical and human factors~\cite{ferryman2023considering}. This includes designing systems that minimize potential technology- induced disparities, taking into account the data and algorithmic literacy of the users of these systems, i.e., clinicians. 
Arbitrariness is not a new concept in the health domain, as evidenced by the existence of cost-effective pharmacological treatments that exhibit suboptimal efficacy in particular patient subgroups. Systems are in place to educate and safeguard against potential patient harm, including rigorous and multiphase pharmaceutical clinical studies and pharmacovigilance protocols.
Data quality audits should scrutinize performance differentials impacting specific subgroups, whose data characteristics may differ from other populations with the same condition.
Future research could explore the creation of monitoring processes for ML models in healthcare, analogous to those applied to pharmacological drugs.

\section*{Conclusion}
In this study, we identified significant age-related and mild sex-related disparities in the performance of ML models for chronic disease prediction. Older patients, in particular, experienced inconsistent and arbitrary predictions across several datasets due to increased data complexity and lower model performance, while sex-based differences slightly favored male predictions. These findings demonstrate that representativeness in training data alone is insufficient for ensuring equitable outcomes. Therefore, addressing model arbitrariness, especially for older individuals, is essential before deploying ML models in clinical settings to ensure fairness and reliability.

\subparagraph{Acknowledgments}
This work is funded by Spanish Ministry of Science and Innovation (REF:DIN2021-011865).
In addition, it has been partially supported by the Department of Research and Universities of the Government of Catalonia (SGR 00930), EU-funded project "SoBigData++" (grant agreement 871042), and MCIN/AEI /10.13039/501100011033 under the Maria de Maeztu Units of Excellence Programme (CEX2021-001195-M). None of the funders played any role in study design, data collection, analysis and interpretation of data, or the writing of this manuscript.

%%%%%%%%%%%%%%%%%%%%%%%%%%%%%%%%%%%%%%%%%%%%%%%%%%%%%%%%%%%%%%%%
\subparagraph{Ethics and Impact Statement}
This study did not involve new data collection; all accessed and processed data were anonymized in compliance with ethical standards and protecting participant privacy. 

%%%%%%%%%%%%%%%%%%%%%%%%%%%%%%%%%%%%%%%%%%%%%%%%%%%%%%%%%%%%%%%%
\subparagraph{Data Availability}
A version of the datasets used during the current study that has been pre-processed for reproducibility is available in the following repository, including pointers to the original datasets: % can be found in \cite{bilionis2023survey}.
\\
\url{https://github.com/IoannisBil/health_disparities_Chronic_Diseases/blob/main/README.md}

% \paragraph{Sample paragraph heading}\lipsum[1]

% \subparagraph{Acknowledgments}\lipsum[1]

% References as numbers
\makeatletter
\renewcommand{\@biblabel}[1]{\hfill #1.}
\makeatother

% unstr is used to keep citation order
% \section*{References}
\bibliographystyle{vancouver}
\bibliography{references.bib}

%%%%%%%%%%%%%%%%%%%%%%%%%%%%%%%%%%%%%%%%%%%%%%%%%%%%%%%%%%%%%%%%

\end{document}